# Evaluating the performance and fragility of large language models on the self-assessment for neurological surgeons

Krithik Vishwanath[1,3,4], Anton Alyakin[1,7], Mrigayu Ghosh[5,6], Jin Vivian Lee[1,7], Daniel Alexander Alber[1], Karl L. Sangwon[1], Douglas Kondziolka[1], Eric Karl Oermann[1,2,8]

[1]Department of Neurological Surgery, [2]Department of Radiology, NYU Langone Medical Center, New York, New York, USA
[3]Aerospace Engineering & Engineering Mechanics, [4]Mathematics, [5]Biomedical Engineering, [6]Molecular Biosciences, The University of Texas at Austin, Austin, Texas, USA
[7]Department of Neurosurgery, Washington University School of Medicine in St. Louis, St. Louis, Missouri, USA
[8]Center for Data Science, New York University, New York, New York, USA

**Correspondence:**
Krithik Vishwanath
Department of Neurosurgery,
NYU Langone Medical Center,
New York University, 550 First Ave, MS 3 205,
New York, NY10016, USA.
Email: krithik.vish@utexas.edu

Eric K. Oermann, MD
Department of Neurosurgery,
NYU Langone Medical Center,
New York University, 550 First Ave, MS 3 205,
New York, NY10016, USA.
Email: eric.oermann@nyulangone.org

**Funding:** EKO is supported by the National Cancer Institute's Early Surgeon Scientist Program (3P30CA016087-41S1) and the W.M. Keck Foundation.

**Disclosures:** EKO has equity in Delvi, MarchAI, and Artisight, income from Merck & Co. and Mirati Therapeutics, employment in Eikon Therapeutics, and consulting for Sofinnova Partners and Google. The other authors have no personal, financial, or institutional interest pertinent to this article.

**Acknowledgements:** We would like to acknowledge Nader Mherabi and Dafna Bar-Sagi, Ph.D., for their continued support of medical AI research at NYU. We thank Michael Constantino, Kevin Yie, and the NYU Langone High-Performance Computing (HPC) Team for supporting computing resources fundamental to our work.

**ABSTRACT**

**BACKGROUND AND OBJECTIVES:** The Congress of Neurological Surgeons Self-Assessment for Neurological Surgeons (CNS-SANS) questions are widely used by neurosurgical residents to prepare for written board examinations. Recently, these questions have also served as benchmarks for evaluating large language models' (LLMs) neurosurgical knowledge. LLMs show significant promise for transforming neurosurgical practice; however, they are susceptible to in-text distractions and confounding factors. Given the increasing use of generative AI and ambient dictation technologies, clinical text is at a larger risk for the inclusion of extraneous details. This study aims to assess the performance of state-of-the-art LLMs on neurosurgery board-like questions and to evaluate their robustness to the inclusion of distractor statements.

**METHODS:** A comprehensive evaluation was conducted using 28 state-of-the-art large language models. These models were tested on 2,904 neurosurgery board examination questions derived from the CNS-SANS. Additionally, the study introduced a distraction framework to assess the fragility of these models. The framework incorporated simple, irrelevant distractor statements containing polysemous words with clinical meanings used in non-clinical contexts to determine the extent to which such distractions degrade model performance on standard medical benchmarks.

**RESULTS:** 6 of the 28 tested LLMs achieved board-passing outcomes, with the top-performing models scoring over 15.7% above the passing threshold. When exposed to distractions, accuracy across various model architectures was significantly reduced—by as much as 20.4% - with one model failing that had previously passed. Both general-purpose and medical open-source models experienced greater performance declines compared to proprietary variants when subjected to the added distractors.

**CONCLUSIONS:** While current LLMs demonstrate an impressive ability to answer neurosurgery board-like exam questions, their performance is markedly vulnerable to extraneous, distracting information. These findings underscore the critical need for developing novel mitigation strategies aimed at bolstering LLM resilience against in-text distractions, particularly for safe and effective clinical deployment.

**Running Title:** LLMs on SANS

**Introduction**

Large language models (LLMs) are rapidly entering clinical neurosurgery, with growing applications in imaging interpretation, intraoperative decision-making, documentation workflows, and education.[1-5] The Congress of Neurological Surgeons' Self-Assessment in Neurological Surgery (CNS-SANS) question bank – long regarded as a gold standard for both resident training and formal assessment— has recently emerged as a benchmark for evaluating whether LLMs can perform at or above the level of a neurosurgical resident in clinical knowledge.[6-11] Recent studies show that leading LLMs such as GPT-3.5 and GPT-4 can exceed board-passing thresholds on neurosurgery[6,7] and general medicine (USMLE) boards-style questions[12,13]. Further, even open-source, light-weight language models that can run on a phone, such as MedMobile, perform well across multiple medical QA benchmarks[14]. Collectively, these findings suggest that LLMs may soon be viable tools for clinical education and point-of-care decision support in neurosurgery.

However, these benchmarks typically involve clean, highly structured prompts that lack the ambiguity and noise of real-world clinical documentation—where tangential information, inconsistently phrased findings, and incomplete data are common. In practice, neurosurgeons must filter noise to identify critical facts, such as a subtle neurological sign or a relevant imaging finding, that inform life-altering decisions. Unlike human clinicians, who are trained to filter irrelevant content, LLMs appear highly susceptible to noisy or distracting inputs. Omar et al.[15] demonstrated that embedding fabricated clinical facts into prompts led to confident hallucinations in over 50–80% of cases. Vishwanath et al.[16] found that inserting irrelevant distractors, such as polysemous terms used in non-clinical contexts, into clinical MCQ questions decreased accuracy by up to 18%, and that fine-tuning or retrieval augmentation failed to mitigate this effect. Ness et al.[17] confirmed that even subtle changes to medical prompts can flip correct answers to incorrect ones — mistakes that trained clinicians would not make.

Despite growing enthusiasm for LLMs in neurosurgery, their ability to deal with noisy or adversarial input remains poorly understood. This is especially pertinent as real-world clinical scenarios are replete with extraneous details – patient histories often include tangential information, and not every detail in a case presentation is pertinent to the diagnosis or management question at hand. These vulnerabilities are especially concerning amid the rapid deployment of ambient AI documentation tools in healthcare. Systems like AI scribes transcribe patient encounters and draft clinical notes in real time, promising to reduce clinician administrative burden and streamline workflows.[18–20] However, they also introduce extraneous or poorly contextualized information into the clinical record — raising the risk that downstream LLMs will misinterpret or overemphasize non-essential or misleading content. In high-stakes fields like neurosurgery, where precision is paramount, an LLM misled by noise could produce unsafe recommendations with direct implications for patient safety.

To address these gaps, we make two primary contributions. First, we evaluate 28 LLMs — spanning open-source, proprietary, general-purpose, and medically fine-tuned models — on 2,904 CNS-SANS board-style questions across all neurosurgical subspecialties. Second, we evaluate model resilience to distractions by systematically injecting clinically irrelevant distractors into prompts. This allows us to simulate the complexity of real-world documentation and assess how such noise degrades model performance. Our goal is to benchmark accuracy and identify the conditions under which LLMs fail – and ultimately, to elucidate the clinical risks of those failures as neurosurgery moves toward AI integration.

**Methods**

<u>Pre-approvals and ethics</u>
The Congress of Neurological Surgeons approved the use of SANS questions in this study. No IRB approval was sought as this study did not involve human subjects. No procedures were performed as a part of this artificial intelligence model benchmarking study.

<u>SANS benchmark</u>
Our primary benchmark was derived from the SANS questions. Developed by the CNS, these questions are a standard resource for neurosurgery residents preparing for the examinations conducted by the American Board of Neurological Surgeons. The CNS provided a total of 3,965 questions for our evaluation. Of these, 1,061 questions included supplementary assets, such as images, and were excluded, leaving 2,904 text-only questions as our benchmark. Evaluation on the CNS-SANS test bank was conducted with explicit permission from the CNS.

SANS with distraction benchmark curation

We began by isolating clinically relevant terms from the incorrect answer options of each SANS item. By selecting only distractor terms, rather than those associated with the correct response, we craft content lacking clinical utility. Specifically, we parsed each question's list of wrong choices and extracted medical concepts (e.g., diseases, conditions, or procedures). For example, if "cancer" appeared among the distractors for a given question, we designated that phrase for possible inclusion in a non-clinical context. This approach exploits the natural variety of erroneous options, often featuring commonplace medical terminology amenable to reinterpretation.

Next, we employed GPT-4o to convert each extracted term into a concise, coherent sentence that utilizes the medical word in a socially or culturally oriented—but clinically irrelevant—context. Prompts instructed the model to generate statements that read fluidly yet convey no diagnostic or therapeutic information. For instance, we might add in the sentence "The patient's zodiac sign is Cancer," in which "Cancer" refers to the astrological sign rather than the malignancy. These confounding sentences were then interwoven into the original SANS questions, yielding augmented items with noise. By systematically introducing irrelevant medical language, we assessed the impact of such distractions on the model's ability to select the correct clinical answer. The overall workflow is illustrated in **Fig. 1**.

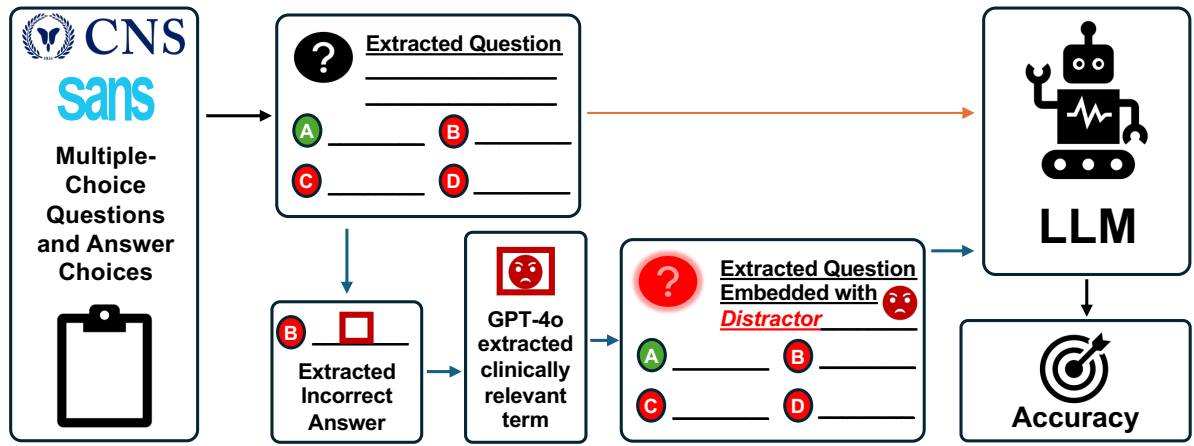

**Figure 1.** Schematic of our study. Problems from the SANS were extracted and fed into LLMs and evaluated for their accuracy in answering the questions correctly (black-orange arrows). The same extracted problems were also used to extract an incorrect answer choice, extract a clinically relevant term from that incorrect answer choice using GPT-4o, and algorithmically embed the clinically relevant term from the wrong answer choice into the original question. The modified problems, including the questions with "distractors," were then similarly fed into the LLMs and evaluated for accuracy (black-blue arrows).

Model evaluation

Accuracy on the SANS and SANS-Noisy benchmarks was assessed by applying exact string matching to each model's chain-of-thought output. All models were run with deterministic decoding—i.e., temperature set to zero—and without ensembling, unless the model inherently disallows temperature control. We employed the providers' default inference parameters in such cases (for example, the OpenAI reasoning models). All open-source and medically fine-tuned models were served using the *PyTorch* package[21], the Hugging Face *Transformers* library[22], and the *vLLM* inference library[23] on identical A100 hardware to ensure a fair comparison. Inference from proprietary systems was obtained through their official APIs, with each call executed according to the vendor's recommended usage guidelines.

Prompting

Prompting templates and prompt engineering for model evaluations follows the method detailed in Vishwanath et al[15].

Categorization of CNS-SANS questions

For each question in the CNS-SANS test bank, we employ a prompting scheme (**see Supplemental Fig. 1**) to identify its theme from 12 official categories of the neurosurgery board exam: Peripheral Nerve, Functional, Fundamentals, Spine, Pain, Tumor, Pediatrics, Trauma, Vascular, Neuropathology, Neuroradiology, and Other General.

Statistical analysis

All statistical analysis was carried out in GraphPad Prism 10.4.2 (Dotmatics) with significance determined at P < .05. ANOVA with Tukey post-hoc analysis was conducted to reveal any significant differences between LLM accuracies. The Spearman's rank correlation coefficient was calculated and tested for any significant associations between rankings. All results are reported as the mean ± standard error of the mean unless otherwise specified.

**Results**

Distractors' impact on SANS accuracy

Once we input the unmodified questions and the questions with distractors into the LLMs, the accuracies of each model were evaluated and subsequently analyzed (**Fig. 1**). Regardless of the model, the presence of distractors in the questions led to a decrease in the accuracy of the LLMs (**Fig. 2A**). However, we observed that the reductions were not symmetrical across the different model classifications of either general open-source; medical open-source; or proprietary.

Considering just the baseline performance, the proprietary models significantly outperformed both open-source models (P = .0280 compared to medical open-source; P < $10^{-4}$ compared to general open-source; **Fig. 2B**). Interestingly, the same trend held for the performance with distractors present within the questions (P = .0188 comparing proprietary to medical open-source; $P < 10^{-4}$ comparing proprietary to general open-source; **Fig. 2C**). Nonetheless, when considering the loss in accuracy between the baseline performance and performance with distractors, the proprietary models significantly outperformed only the general open-source models (P = .0001; **Fig. 2D**). This suggests that the proprietary models are not only the most robust in the baseline case and with distractors present, but also the most resilient to the addition of distractors in the prompt. On the other hand, general open-source models are the least resilient to the addition of distractors.

We confirmed this trend by conducting a spearman correlation analysis between the baseline performance and change in performance with distractors. Our analysis revealed a moderately strong negative correlation between the baseline accuracy and the loss in accuracy with distractors present ($\rho$ = -0.66, P = .0001; **see Supplemental Fig. 2**). Additionally, general open-source models clustered around the high accuracy loss and low baseline accuracy while proprietary models clustered around the low accuracy loss and high baseline accuracy. This further highlights the robustness and resilience of proprietary LLMs to added distractors.

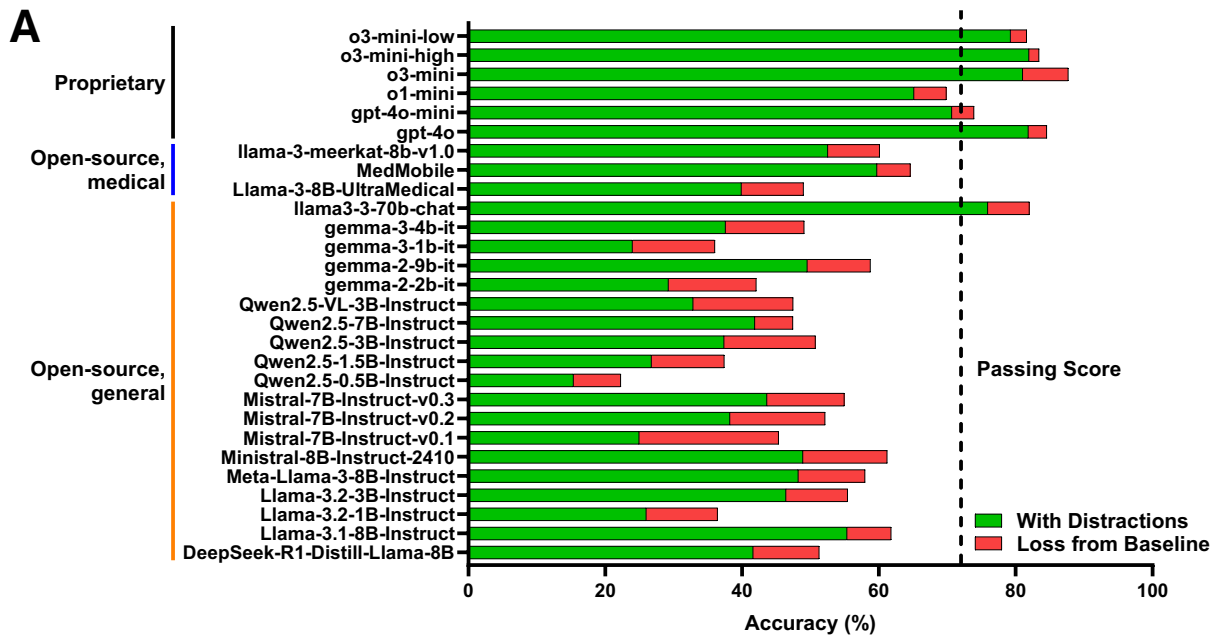

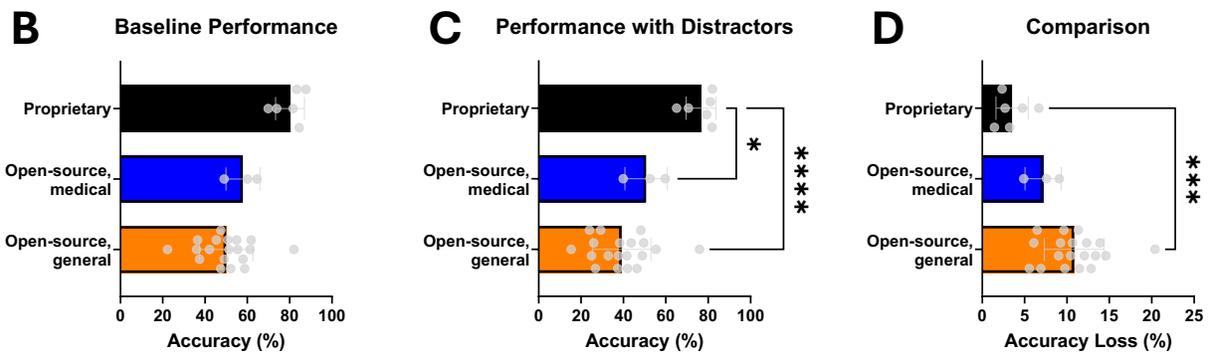

**Figure 2.** Accuracy of LLMs in answering SANS questions with and without distractors present by model classification. **A**, Accuracy of LLMs in answering SANS questions with distractors (green) and the respective loss from the baseline accuracy without any distractors (red). The baseline accuracy of a given LLM is the respective total length of the bar (green plus red). Models are grouped by their classifications as general open-source (orange); medical open-source (blue); and proprietary (black). **B**, Baseline accuracy of general open-source; medical open-source; and proprietary LLMs in predicting answers to SANS questions without any distractors. **C**, Accuracy of general open-source; medical open-source; and proprietary LLMs in predicting answers to SANS questions with distractors present. **D**, Loss in accuracy of general open-source; medical open-source; and proprietary LLMs from the baseline accuracy to the accuracy with distractors present. Bars indicate significant differences ($P < 0.05$ one-way ANOVA with Tukey post-hoc analysis, $*P < 0.05$, $***P < 0.001$, $****P < 0.0001$; general open-source, $n = 19$; medical open-source, $n = 3$; proprietary, $n = 6$).

Distractor impact on SANS by category

After discovering the model classification dependence of the fragility of LLMs to distractors, we analyzed whether the LLMs' performance differed by the section of the SANS exam from which the question came. Without any distractors present, the LLMs performed the most accurately in the Fundamentals, Neuropathology, and Other General sections ($P < .0049$; **Fig. 3A**). With distractors present, though, Neuropathology was no longer among the most accurately answered sections, and only Fundamentals and Other General remained among the most accurately answered sections ($P < .0128$; **Fig. 3B**). The same sections—Functional, Spine, Vascular, and Neuroradiology—were consistently the lowest-performing, regardless of the presence of distractors.

We analyzed the loss in accuracy between the baseline performance and the performance with distractors, both by section and model classification. As expected, for nearly all the sections, the proprietary models exhibited a significantly lower accuracy loss than general open-source models ($P < .0330$ for all sections except Functional for which $P = .0952$, Spine for which $P = .3139$, and Trauma for which $P = .4649$; **Fig. 3C**).

Additionally, we observed that the accuracy loss for Neuropathology was significantly higher than that of other sections only for the general open-source models ($P < 10^{-4}$ compared to Functional, $P = .0069$ compared to Fundamentals, $P < 10^{-4}$ compared to Spine, $P = .0463$ compared to Pain, $P < 10^{-4}$ compared to Trauma, $P < 10^{-4}$ compared to Vascular for open-source general; $P > .3015$ for open-source medical; $P > .9975$ for proprietary; **Fig. 3C**). Thus, the drop in accuracy of LLMs answering Neuropathology questions can primarily be attributed to the higher accuracy loss for Neuropathology questions in general open-source models.

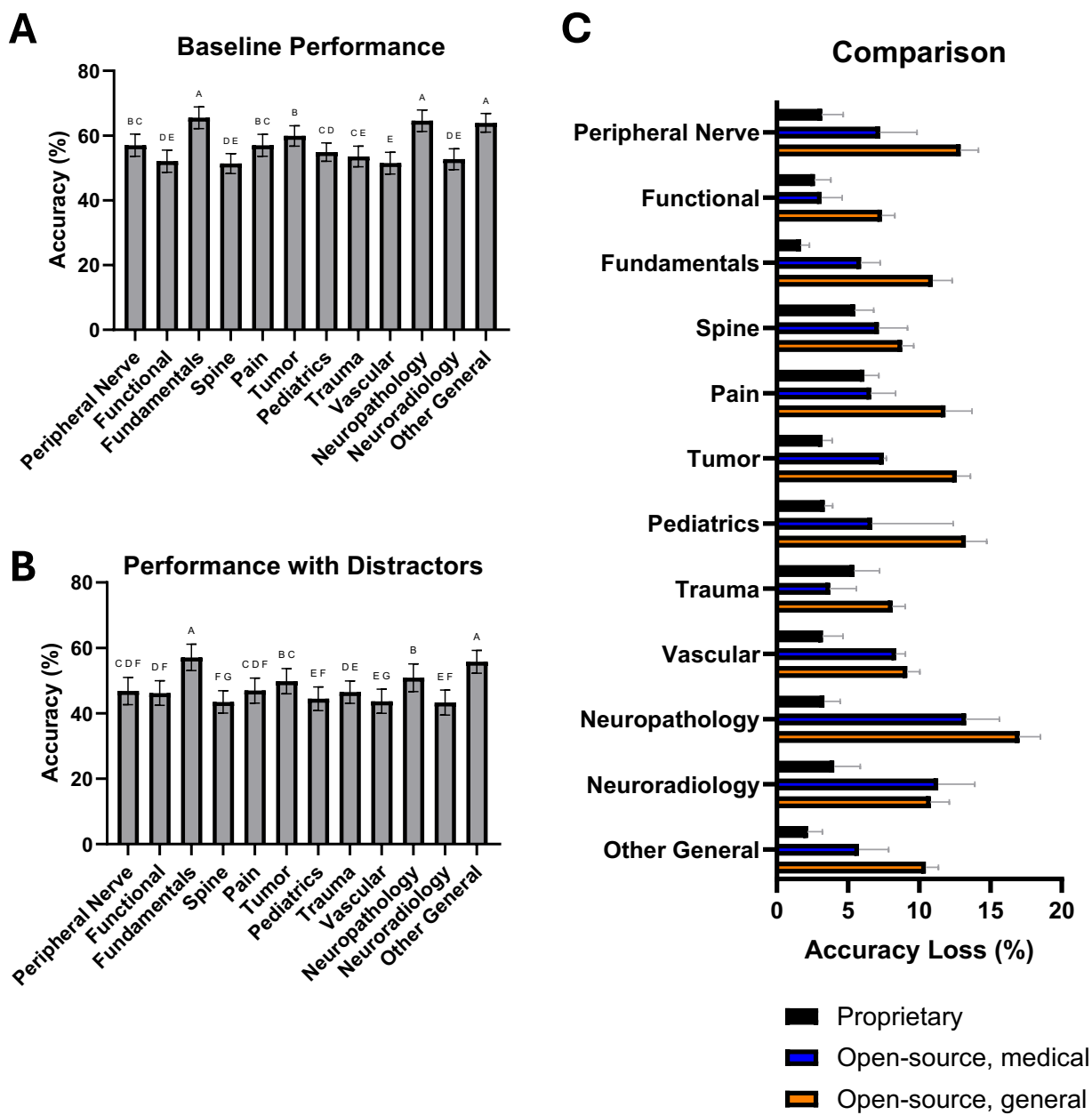

**Figure 3.** Accuracy of LLMs in answering SANS questions with and without distractors present by question section. **A**, Baseline accuracy of LLMs in predicting answers to SANS questions of each section without any distractors. **B**, Accuracy of LLMs in predicting answers to SANS questions of each section with distractors present. Different letters indicate significantly different values ($P < 0.05$, one-way ANOVA with Tukey post-hoc analysis; $n = 28$). **C**, Loss in accuracy of LLMs from the baseline accuracy to the accuracy with distractors present, by model classification and by section.

**Discussion**

As LLMs continue to demonstrate remarkable capabilities in medical knowledge recall and board question-answering, interest in their application within neurosurgical training and clinical decision support has surged. From an evaluation of 2904 SANS questions across 28 LLMs, our results confirm this ability of LLMs, both proprietary and open-source, to encode clinical neurosurgical knowledge but also reveal a previously underappreciated vulnerability: when exposed to irrelevant yet clinically familiar distractors, LLMs often exhibit substantial degradation in performance (up to 20.4%), with potentially grave implications for both neurosurgical education and patient care. The safe and effective integration of LLMs into neurosurgical workflows requires a nuanced understanding of their limitations, particularly in high-stakes environments where cognitive precision and distraction-filtering are essential.

By systematically embedding non-informative, medically adjacent language into board-style neurosurgical questions, we observed that all tested models—regardless of architecture or training domain—experienced measurable declines in accuracy. This effect was most pronounced among general open-source-purpose LLMs, which not only demonstrated the lowest baseline performance but also the greatest susceptibility to distraction (5.6% to 20.4% drop in accuracy, depending on the specific model). Proprietary models, while more resilient, were not immune to these effects (for comparison, 1.4% to 6.7% drop in accuracy, depending on the specific model). These findings suggest that current LLMs lack the contextual filtering capabilities that are second nature to trained neurosurgeons.

From an educational standpoint, these results are particularly salient. The growing use of LLMs for resident training, simulated board examinations, and just-in-time learning assessments assumes that models can correctly parse clinically relevant content. Our findings are consistent with prior research by Omar et al.[14] and Vishwanath et al.[15], who report that even top-performing models can be misled by semantically plausible but irrelevant information. In the context of neurosurgical education, such errors could erode trainee confidence, reinforce misconceptions, or misdirect study focus—particularly when models are trusted to provide reasoning or explanations alongside answers.

Clinically, the distractibility of LLMs raises concerns about patient safety. As LLMs are increasingly embedded into electronic health record (EHR) systems for ambient documentation, summarization, and even clinical decision support[19,20], their ability to prioritize relevant information becomes essential. Medical documentation often contains redundant, templated, or copy-forwarded text—elements that a human clinician may intuitively disregard but which could confuse a model. This vulnerability is compounded in high-volume, cognitively intense environments such as the emergency department, trauma bay, or neurosurgical ICU, where clarity and concision are paramount. Such errors, especially when unchecked, have grave implications for clinical decision-making and patient safety.

From a technical standpoint, our study underscores the need for robust evaluations that reveal insights about LLMs beyond benchmark accuracy. Recent work has shown that LLMs can fail under subtle perturbations[17], variation in clinical note type[24], training data-poisoning[25], or even internal tampering[26], and that fine-tuning alone does not confer resistance to distractions[15]. Although medical fine-tuning improves content familiarity, it does not reliably teach models to suppress misleading or non-salient information—a critical feature for clinical deployment. To this end, future systems may require technical and architectural innovations such as adversarial noise-aware fine-tuning[27,28] and retrieval-augmented prompting restricted to curated templates[29,30].

Importantly, our analysis revealed domain-specific susceptibility: questions from the Neuropathology section were disproportionately affected by distractors, especially in open-source models. This may reflect the abstract, clinically inconsequent, and often polysemous language used in neuropathology, which challenges even experienced clinicians. In contrast, sections such as Fundamentals and Other General were more robust, suggesting that the semantic clarity of content plays a key role in LLM resilience.

Limitations

Several factors may temper the generalizability of these findings. First, roughly one-quarter of the original SANS corpus—those items containing images—was excluded; multimodal LLMs may yield different robustness profiles in the SANS dataset as demonstrated in Alyakin et al[1]. Second, only a single distractor style was examined, whereas real-world clinical notes may feature longer digressions, copy-and-paste redundancy and transcription errors. Third, the SANS-CNS dataset, while representative of the neurological board exams, are not completely analogous to the various properties of a real-world clinical environment and tasks required by physicians. Moreover, multiple-choice formats may inadequately assess nuanced clinical reasoning, as LLM performance on such benchmarks can be driven by selection biases and format-specific artifacts, potentially overestimating their practical utility.[31,32] Finally, exact string-matching evaluation cannot award partial credit for conceptually correct but non-identical answers, potentially underestimating practical competence.

Future work should explore training-time interventions to enhance attention to clinically relevant content, the development of domain-specific safety filters, and real-time monitoring tools to detect LLM uncertainty or confusion. In educational settings, supervised feedback and adversarial testing may help refine model outputs and prevent propagation of errors. As neurosurgery considers the adoption of AI tools, the standard for robustness must match the cognitive rigor of the specialty itself.

**Conclusion**

LLMs have reached and, in some cases, surpassed expert-level performance on clean neurosurgical board questions, but that competence is fragile. A single irrelevant sentence containing a medical term can erode accuracy by up to 20.4%, with open-source general-purpose models suffering the most significant declines and proprietary models showing—but not eliminating—residual vulnerability. Because real-world clinical text is replete with extraneous information, robustness to noise is not a peripheral nicety; it is a prerequisite for safe deployment. We therefore advocate that future neurosurgical and broader medical AI benchmarks incorporate explicit noise stress tests and that model developers prioritize training and architectural strategies that confer resilience. Only by demonstrating both high accuracy and high robustness can LLM-based tools earn the trust required for integration into neurosurgical decision making and, ultimately, improve patient care.

## Author Contributions

AA and EKO supervised the study. KV, AA, and EKO conceptualized and established the study design. KV designed and developed the LLM evaluation pipeline and the SANS with distractions benchmark. KV and MG wrote the initial draft and developed the figures of the manuscript. All authors revised and approved the manuscript.

## Data availability

Due to the proprietary nature of the SANS dataset, the datasets generated or analyzed during the current study will NOT be released.

## Code availability

Our code is available via request to the corresponding author.

**Supplemental Digital File**

---

**Prompt Template: `prompt_select_SANS_category`**

## Question:
{{question}}

## Choices:
{{choices}}

1. Peripheral Nerve
2. Functional
3. Fundamentals
4. Spine
5. Pain
6. Tumor
7. Pediatrics
8. Trauma
9. Vascular
10. Neuropathology
11. Neuroradiology
12. Other General

Only respond with the topic number (1-12) that BEST fits with the medical question. Do not include any other text in your response.

---

**Supplemental Digital Content 1, Figure.** Prompting template for categorizing questions within neurosurgical subtopics.

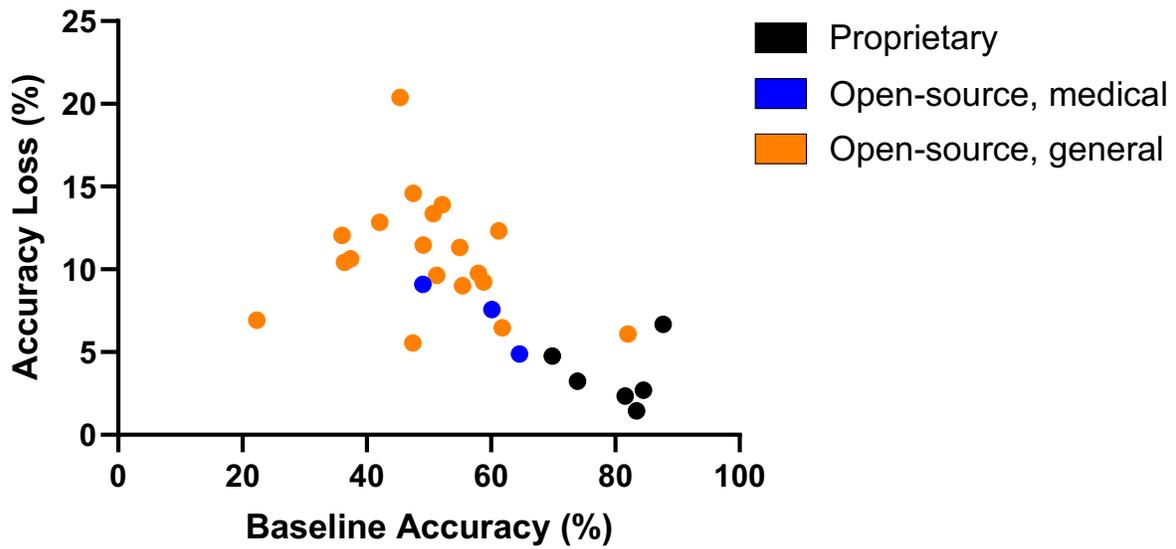

**Supplemental Digital Content 2, Figure.** Correlation between baseline accuracy and accuracy loss under distractors present of each LLM by model classification. Each circle represents a different LLM, and points are colored by model classification (orange: general open-source; blue: medical open-source; black: proprietary). The Spearman correlation coefficient is $\rho$ = -.66 with $P$ = .0001.

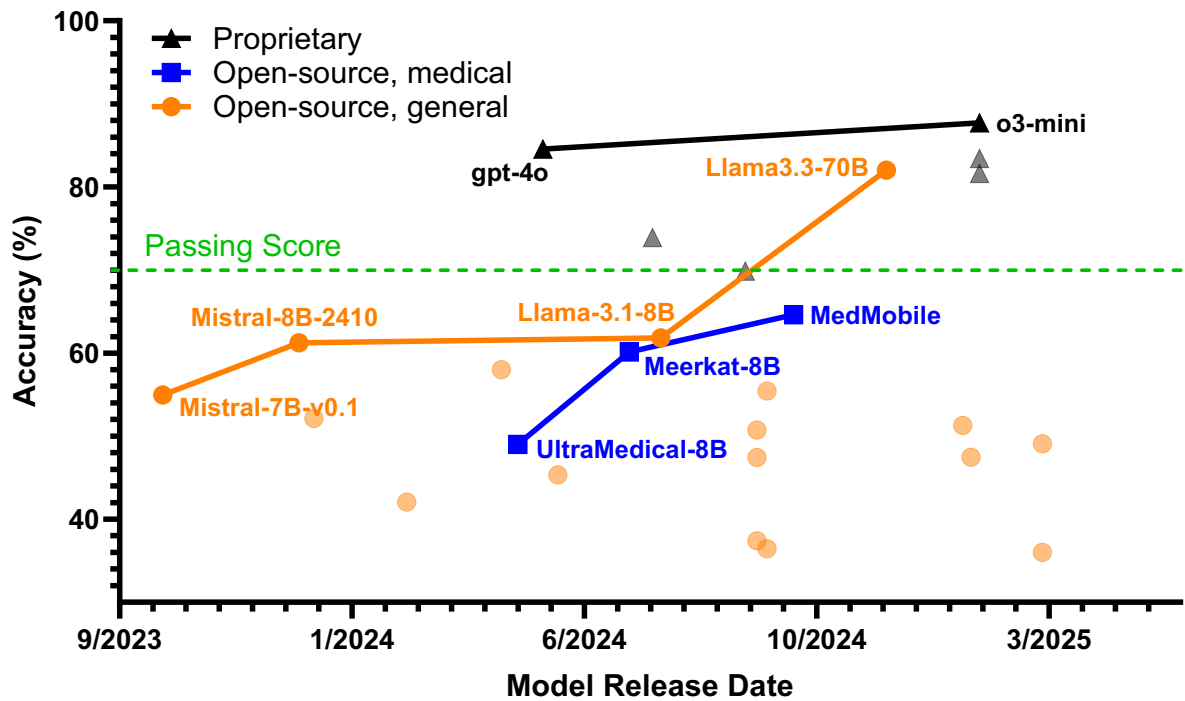

**Supplemental Digital Content 3, Figure.** Timeline progression of LLM performance on the neurosurgical board-like exam questions. Each marker represents a different LLM, and points are colored by model classification (orange: general open-source; blue: medical open-source; black: proprietary). Connecting lines are included to show improvement in the baseline performance over time by model classification.